\documentclass[times, twoside]{zHenriquesLab-StyleBioRxiv}

\leadauthor{B\"uchel} 

\usepackage{url,lineno,microtype}

\usepackage{tabularx, multirow}
\usepackage{makecell}
\usepackage{booktabs}
\usepackage{wrapfig}

\usepackage[acronym]{glossaries}
\newacronym{cnn}{CNN}{Convolutional Neural Network}
\newacronym{snn}{SNN}{Spiking Neural Network}
\newacronym{ann}{ANN}{Artificial Neural Network}
\newacronym{scnn}{SCNN}{Spiking Convolutional Neural Network}
\newacronym{pgd}{PGD}{Projected Gradient Descent}
\newacronym{bptt}{BPTT}{Backpropagation Through Time}
\newacronym{dvs}{DVS}{Dynamic Vision Sensor}
\newacronym{nmnist}{N-MNIST}{Neuromorphic MNIST}

\def\aBioRXiv{{\footnotesize
    \footerfont B\"uchel \textit{et al.}\hspace{7pt}|\hspace{7pt}2022\hspace{7pt}|\hspace{7pt}\thepage\textendash\pageref{LastPage}
  }}

\def\bBioRXiv{{\footnotesize
    \footerfont \thepage\hspace{7pt}
  }}
  
\def\cBioRXiv{\hspace{7pt}\thepage}

\begin{document}

\title{Adversarial Attacks on Spiking Convolutional Neural Networks for Event-based Vision}
\shorttitle{Adversarial Attacks on SCNNs for Event-based Vision}

\author[1]{Julian B\"uchel}
\author[2]{Gregor Lenz}
\author[3]{Yalun Hu}
\author[2]{Sadique Sheik}
\author[4]{Martino Sorbaro}

\affil[1]{IBM Research, Z\"urich, Switzerland}
\affil[2]{SynSense, Z\"urich, Switzerland}
\affil[3]{SynSense, Chengdu, China}
\affil[4]{AI Center, ETH Z\"urich and Institute of Neuroinformatics, University of Z\"urich and ETH Z\"urich, Switzerland}

\maketitle



\begin{corrauthor}
jub@zurich.ibm.com, msorbaro@ethz.ch
\end{corrauthor}

\begin{abstract}
Event-based dynamic vision sensors provide very sparse output in the form of spikes, which makes them suitable for low-power applications. Convolutional spiking neural networks model such event-based data and develop their full energy-saving potential when deployed on asynchronous neuromorphic hardware. Event-based vision being a nascent field, the sensitivity of spiking neural networks to potentially malicious adversarial attacks has received little attention so far. We show how white-box adversarial attack algorithms can be adapted to the discrete and sparse nature of event-based visual data, and demonstrate smaller perturbation magnitudes at higher success rates than the current state-of-the-art algorithms. For the first time, we also verify the effectiveness of these perturbations directly on neuromorphic hardware. Finally, we discuss the properties of the resulting perturbations, the effect of adversarial training as a defense strategy, and future directions.
\end{abstract}

\maketitle

\section*{Introduction}
Compared to the neural networks commonly used in deep learning, \glspl{snn} resemble the animal brain more closely in at least two main aspects: the way their neurons communicate through spikes, and their dynamics, which evolve in continuous time. Aside from offering more biologically plausible neuron models for computational neuroscience, research in the applications of \glspl{snn} is currently blooming because of the rise of neuromorphic technology. Neuromorphic hardware is directly compatible with \glspl{snn} and enables the design of low-power models for use in battery-operated, always-on devices.

Adversarial examples are an ``intriguing property of neural networks'' \citep{szegedy2013intriguing} by which the network is easily fooled into misclassifying an input which has been altered in an almost imperceptible way by the attacker. This property is usually undesirable in applications: it was proven, for example, that an adversarial attack may pose a threat to self-driving cars \citep{eykholt2018robust}. Because of their relevance to real-world applications, a large amount of work has been published on this subject, typically following a pattern where new attacks are discovered, followed by new defense strategies, in turn followed by proof of other strategies that can still break through them (see \citealt{adv-review} for a review).

With the advent of real-world applications of spiking networks in neuromorphic devices, it is essential to make sure they work securely and reliably in a variety of contexts. In particular, there is a significant need for research on the possibility of adversarial attacks on neuromorphic hardware used for computer vision tasks.
In this paper, we make an attempt at modifying event-based data, by adding and removing events, to generate adversarial examples that fool a spiking network deployed on a convolutional neuromorphic chip. This offers important insight into the reliability and security of neuromorphic vision devices, with important implications for commercial applications.

\subsection{What is event-based sensing?}
Event-based \glspl{dvs} share characteristics with the mammalian retina and have several advantages over conventional, frame-based cameras: 
\begin{itemize}
  \item Camera output in the form of events and thus power consumption are directly driven by changes in the visual scene, omitting output completely in the case of a static scene.
  \item Pixels fire independently of each other which results in a stream of events at microsecond resolution instead of frames at fixed intervals. This enables very low latency and high dynamic range.
\end{itemize}

The sparse, asynchronous \gls{dvs} output does not suit current high-throughput, synchronous accelerators such as GPUs. To process event-based data efficiently, neuromorphic hardware is being developed, where neurons are only updated whenever they receive an event. 
Spiking neuromorphic systems include large-scale simulation of neuronal networks for neuroscience research \citep{furber2012overview} and low-power real-world deployments of machine learning algorithms. \glspl{scnn} as well as conventional \glspl{cnn} have been run on neuromorphic chips such as IBM's TrueNorth and HERMES \citep{esser2016convolutional,hermes}, Intel's Loihi \citep{davies2018loihi} and SynSense's  Speck and Dynap-CNN \citep{liu2019live} for low-power inference. The full pipeline of event-based sensors, stateful spiking neural networks, and asynchronous hardware -- which is present in SynSense's Speck -- allows for large gains in power efficiency compared to conventional systems. 

\subsection{Adversarial attacks on discrete data}
The history of attack strategies against various kinds of machine learning algorithms pre-dates the advent of deep learning \citep{biggio2018wild}, but the phenomenon received widespread interest when adversarial examples were first found for deep convolutional networks \citep{szegedy2013intriguing}. In general, given a neural network classifier $C$ and an input $x$ which is correctly classified, finding an adversarial perturbation means finding the smallest $\delta$ such that $C(x+\delta) \neq C(x)$. Here, ``smallest'' refers to minimizing $\|\delta\|$, where the norm is chosen arbitrarily depending on the requirements of the experiment. For example, using the $L^\infty$ norm (maximum norm) will generally make the perturbation less noticeable to a human eye, while the use of the $L^1$ norm will encourage sparsity, i.e.\ a smaller number of perturbed pixels.

There are two main challenges in transferring existing adversarial algorithms to event-based vision:  
\begin{itemize}
    \item The presence of a continuous time dimension, as opposed to frames taken at fixed intervals;
    \item The binary discretization of input data and SNN activations, as opposed to traditional image data (at least 8 bit) and floating point network activations. 
\end{itemize}

\begin{figure}
    \centering
    \includegraphics[width=0.9\columnwidth]{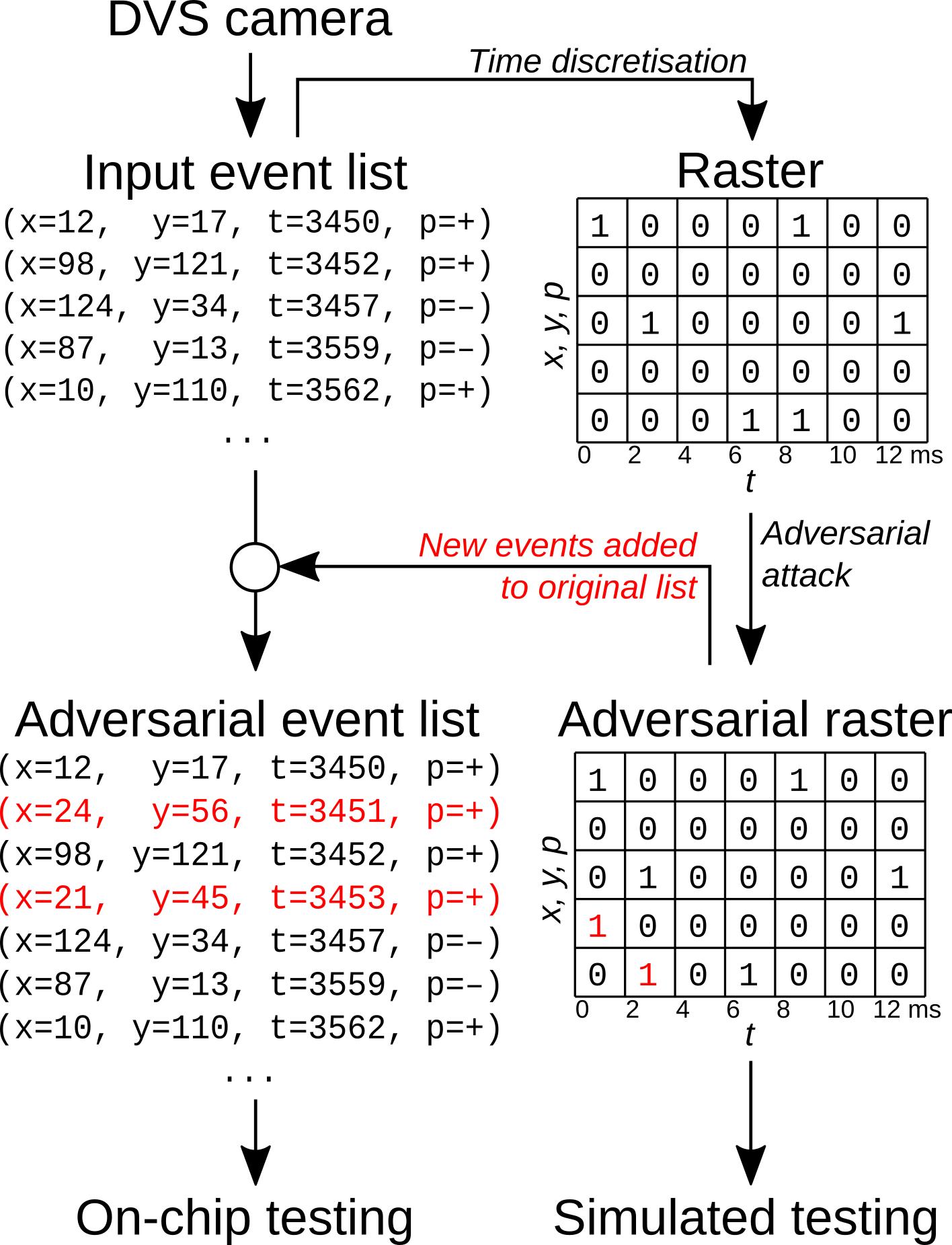}
    \caption{Schematic of the attack procedure on \gls{dvs} data.}
    \label{fig:schema}
\end{figure}

Event-based sensors encode information in the form of events that have a timestamp, location $(x, y)$ and polarity (lighting increased or decreased). Because at any point in time an event can either be triggered or not, one can simply view event-based inputs as binary data by discretizing time (Figure \ref{fig:schema}). In this view, the network's input is a three-dimensional array whose entries describe the number of events at a location $(x,y)$ and in time bin $t$; an additional dimension, of size 2, is added due to the polarity of events. If the time discretization is sufficiently precise, and no more than one event appears in each voxel, the data can be treated as binary.

In this work, we present new algorithms that adapt the adversarial attacks SparseFool~\citep{sparsefool}, and adversarial patches~\citep{patches}, to work with the time dynamics of spiking neural networks, and with the discrete nature of event-based data. We focus on the case of white box attacks, where the attacker has full access to the network and can backpropagate gradients through it. We test our attacks on the \gls{nmnist}~\citep{nmnist} and IBM Gestures~\citep{ibm-gestures} datasets, which are the most common benchmark datasets within the neuromorphic community. Importantly, for the first time, we also test the validity of our methods by deploying the attacks on neuromorphic hardware.

Our contributions can be summarized as follows:
\begin{itemize}
    \item We contribute algorithms that adapt several adversarial attacks strategies to event-based data and \glspl{snn}, with detailed results to quantify their effectiveness and scalability.
    \item We show that these adapted algorithms outperform state-of-the-art algorithms in the domain of \glspl{scnn}.
    \item We show targeted universal attacks on event-based data in the form of adversarial patches, which do not require prior knowledge of the input.
    \item We validate the resulting adversarial examples on an \gls{snn} deployed on a convolutional neuromorphic chip. To the best of our knowledge, this is the first time the effectiveness of adversarial examples is demonstrated directly on neuromorphic hardware.
\end{itemize}

\section*{Related Work} \label{sec:related_work}
Despite the growing number of \glspl{snn} deployed on digital \citep{davies2018loihi} and analog~\citep{dynaps} neuromorphic hardware, robustness to adversarial perturbations has received comparatively little attention by the research community.
Some methods proposed for attacking binary inputs have focused on brute-force searches with heuristics to reduce the search space~\citep{adv-training-prob-snns,scar}. Algorithms of this family do not scale well to large input sizes, as the number of queries made to the network grows exponentially. In particular, this becomes a serious problem when the time dimension is added, greatly increasing the dimensionality of the input.

In \cite{marchisio2021dvs}, the authors demonstrate various algorithms for attacking \gls{dvs} data that serves as input to an \gls{snn}. While some attacks are brute force ("Dash Attack", "Corner Attack") and therefore do not scale, "Frame Attack" simply adds events in hard-coded locations and produces perturbations that are orders of magnitudes larger than all of the perturbations of the algorithms presented here. The only algorithm that exploits gradient information is "Sparse Attack", against which we compare\footnote{We followed the implementation at \url{https://github.com/albertomarchisio/DVS-Attacks}.}. Unfortunately, the authors do not report quantitative results on the magnitudes of their perturbations (i.e. how many spikes are added or removed), which is an important metric for demonstrating efficiency of the attack.

\cite{sharmin2020inherent} demonstrate interesting properties of the inherent robustness of \glspl{snn} to adversarial attacks, with variations depending on the training method. Their work, however, only uses static image data with continuous pixel values converted to Poisson spike input frequencies, which is rather different from working with actual event-based camera input.

As \cite{adv-snn} note, white-box attacks can exploit ``surrogate'' gradients in order to calculate gradients of a loss function with respect to the input in \glspl{snn}. Not only do these gradients reflect the true dependency of the output to the input, but they also capture temporal information via \gls{bptt}. In their work, the authors use a Dirac function for their surrogate gradient, which causes, as they report, considerable ``vanishing gradient'' problems. While they show good results, they were forced to use additional tricks such as probabilistic sampling and an artificial construction of gradients termed ``Restricted Spike Flipper'' in their manuscript. In our work we solved the vanishing gradient issue by resorting to a more effective surrogate function (Section \ref{sec:networks}).

Outside of the domain of \glspl{snn}, various studies explore the adversarial robustness of conventional neural networks deployed on digital accelerators or analog in-memory computing based accelerators. For example, \cite{stutz} demonstrate that standard networks are susceptible to bit errors in the SRAM array storing the quantized weights in deep learning accelerators. The authors further show that one can mitigate this susceptibility via random bit flipping during training. While this work performs experiments using noise models obtained from various SRAM arrays, it still lacks a full-fledged hardware demonstration. \cite{Shimeng} obtain a noise model of a RRAM-based crossbar often found in analog in-memory computing architectures. The noise model is then used in simulation to study the adversarial robustness of neural networks deployed on such a crossbar.

Besides attacking neural networks via input perturbation, a large body of work on directly attacking the physical hardware exists. Hardware fault injection attacks (see \citealt{fia-survey} for a survey) are mostly based on voltage glitching, electromagnetic pulses, or rapid row activation in DRAM~\citep{rowhammer}.

In this work, we focus on input-based attacks, which we verify in simulations and on commercially available spiking neuromorphic hardware. This step is important to determine real-world effectiveness, as the exact simulation of asynchronous, event-based processing with microsecond resolution is prohibitively expensive on conventional hardware.

\section*{Methods} \label{sec:methods}
\subsection{Attack strategies} \label{sec:attacks}
\subsubsection{SparseFool on discrete data}
To operate on event-based data efficiently, the ideal adversarial algorithm requires two main properties: sparsity and scalability. \textit{Scalability} is needed because of the increased dimensionality given by the additional time dimension. \textit{Sparsity} ensures that the number of events added or removed is kept to a minimum.
One approach that combines the above is SparseFool \citep{sparsefool}, which iteratively finds the closest point in $L^2$ on the decision boundary of the network using the DeepFool algorithm \citep{deepfool} as a subroutine, followed by a linear solver that enforces sparsity and boundary constraints on the perturbation. DeepFool finds the smallest perturbation in $L^2$ by iteratively moving the input in the direction orthogonal to the linearized decision boundary around the current input. Since decision boundaries of neural networks are non-linear, this process has to be repeated until a misclassification is triggered.
Because \glspl{snn} have discrete outputs (the number of spikes over time for each output neuron), it is easier to suffer from vanishing gradients as the perturbation approaches the decision boundary. This occurs because DeepFool calculates the perturbation that just reaches the decision boundary, which is very small when the input is already close to the decision boundary. Therefore, adding this small perturbation to the input might not reflect in the number of emitted spikes as the membrane potential must cross the spiking threshold. We made the following changes to overcome these issues. Firstly, we clamped the perturbation at every iteration of DeepFool so that it was no smaller than a value $\eta$, in order to protect against vanishing gradients. $\eta$ was treated as a hyperparameter that should be kept as small as possible without incurring vanishing gradients. Without $\eta$, SparseFool yields a success rate of 12.9\% on 100 samples of the \gls{nmnist} dataset, where the success rate is the percentage of attacked samples that led to a misclassification out of the samples that were originally classified correctly by the network.
Secondly, to account for the discreteness of event-based data, we rounded the output of SparseFool to the nearest integer at each iteration.
Finally, SparseFool normally involves upper and lower bounds $l$ and $u$ on pixel values (normally set, for images, to $l=0;u=255$). We exploit these to enforce the binary constraint on the data ($l=0;u=1$), or, in the on-chip experiments, to fix a maximum firing rate in each time bin, which is the same as that of the original input ($l=0;u=\max(\text{input)}$). From now on, we will refer to this variant as SpikeFool.

\subsubsection{Adversarial patches}
As the name suggests, adversarial patches are perturbations that are accumulated in a certain region (patch) of the image. The idea is that these patches are generated in a way that enables the adversary to place them anywhere in the image. This attack is \textit{targeted} to a desired label, and \textit{universal}, i.e. not input-specific.
To test a more realistic scenario where an adversary could potentially perform an attack without previous knowledge of the input, we apply these patches to the IBM hand gesture dataset.  We note that the prediction of the \gls{cnn} trained on this dataset is mostly determined by spatial location of the input. For example, the original input of ``Right Hand Wave'' is not recognized as such if it is shifted or rotated by a substantial amount. In order to simulate effective realistic attacks, we choose to limit both computed and random attack patches to the area of where the actual gesture is performed.
As in \cite{patches}, we generate the patches using \gls{pgd} on the log softmax value of the target output neuron. \gls{pgd} is performed iteratively on different images of the training set and the position of the patch is randomized after each sample.
For each item in the training data, the algorithm updates the patch until the target label confidence has reached a pre-defined threshold. The algorithm skips the point if the original label equals the target label. This process is repeated for every training sample and for multiple epochs.
To measure the effectiveness of our computed patches, we also generate random patches of the same size, and measure the target success rates. In a random patch, every pixel has a 50\% chance of emitting a spike at each time step.

\subsection{Datasets}
\gls{nmnist} consists of 300 ms-long recordings of MNIST digits that are captured using a three-fold saccadic motion of a \gls{dvs} sensor \citep{nmnist}. We bin the events of each recording into 60 time steps, capping the maximum number of events to 1 per pixel.

IBM Gestures dataset consists of recordings of 11 classes of human gestures, captured under three different lighting conditions \citep{ibm-gestures}. For this dataset, the model must have the ability to process temporal features to distinguish between clockwise and counterclockwise versions of the same gesture. We bin 200\,ms slices of recordings into 20 frames each, again capping the frames to 1 per pixel.
For experiments on the chip we choose a higher temporal resolution and bin the same 200\,ms slices into 100 frames each.

\subsection{Network models} \label{sec:networks}
The spiking networks used for the \gls{nmnist} and IBM Gestures tasks are trained using the publicly available PyTorch-based \gls{snn} library Sinabs\footnote{\url{http://sinabs.ai}} which supports the same non-leaky integrate-and-fire neurons available on the neuromorphic chip. 

Models used for \gls{nmnist} were trained using \emph{weight transfer}, whereby an equivalent \gls{cnn} is trained on accumulated frames (i.e. summing the data over the time dimension), and the \gls{cnn} weights are transferred to the spiking network~\citep{rueckauer2017conversion, sorbaro2020optimizing}. 
The model we used is a LeNet-5 architecture with 20, 32, 128, 500 and 10 channels respectively. 
The network achieves 85.05\% classification test accuracy with full precision weights and 85.2\% with 8-bit quantized weights.

For the IBM Gestures task, training is done using \gls{bptt} since for this dataset we have to learn temporal features as well. 
We make use of a surrogate gradient in the backwards pass to enable learning despite the discontinuous nature of spikes \citep{neftci2019surrogate}.
The model has a LeNet-5 architecture plus batchnorm layers with 8, 8, 8, 64 and 11 channels respectively. 
This network achieves a classification test accuracy of 84.2\%. The accuracy is retained if the network weights are quantized to 8 bits. The network used for the on-chip experiments does not have batch-normalization layers as they introduce biases which are not supported on the hardware. 


\subsection{Experiments on the neuromorphic chip} \label{sec:hardware_methods}
We used a multi-core spiking neural network processing chip prototype from SynSense, called ``Speck2b''. It uses asynchronous logic design to keep dynamic power that is consumed whenever spikes are routed to a minimum. That means that neurons update and spike only when an input is received, but are not limited to time bins of clock cycles, which is very different than conventional von Neumann hardware.
The chip and its 327 000 neurons are fully configurable, supporting convolutional, linear, as well as pooling connectivity. Event input (from datasets such as \gls{nmnist} and IBM gestures) is sent to the FPGA as a block of events in $(t, p, x, y)$ format. The FPGA then handles the feeding to the chip at the correct timestamps. This is important as the chip itself, using fully asynchronous logic, does not understand the notion of time steps. Every neuron will compute its input on demand and fire as soon as it is ready. Since events have to be handled in a sequential manner, this can sometimes lead to slight changes in the order of input events, which then has consequences for neuron activation further downstream (in later layers). If many events arrive in a short time span, they are buffered depending on the throughput limitations of each core. Those limitations together with weight quantization on the neuromorphic hardware lead to slight differences in output when compared to simulation.

The multi-core, single-chip neuromorphic processor we used for the experiments features an integrated DVS sensor for real-time, fully integrated vision and is commercially available for programming as part of a development kit, which includes USB connectivity, power management and an FPGA for communication. Since the chip features an on-chip DVS sensor, its main application is in the vision domain, making it even more relevant for our study.

On the chip, weight precision, number of computations per second and throughput are reduced, as the hardware is optimized for very low power consumption. The networks detailed in the previous sections have to be modified in order to make them suitable for on-chip inference: their weights are rescaled and discretized as required by the chip's 8-bit weight precision. This is one factor that can lead to a degradation in prediction accuracy when compared to simulations. A second and more important factor why simulations do not mimic our chip's behavior exactly is the need for time discretization when training our networks off-chip (see Figure~\ref{fig:schema}).

As this work focuses on white-box attacks, we first compute the adversarial examples using the network simulation on the computer, and then test both original and attacked spiketrains in simulation and on the chip. However, the simulation and the attack work in discrete time, while the chip receives events in continuous time. In order to convert the discrete-time attacked raster back to a list of events for the chip, we compare the original and attacked rasters, identifying new events added by the attack and adding them to the original list (Figure \ref{fig:schema}). We empirically found very few events were removed by SparseFool; for example, in the \gls{nmnist} experiment, there were 28 additions per each removal (see supplementary material for more information). For simplicity, we therefore chose to ignore removals for on-chip experiments.

\section*{Results}

\begin{figure*}
    \centering\includegraphics[width=.9\textwidth]{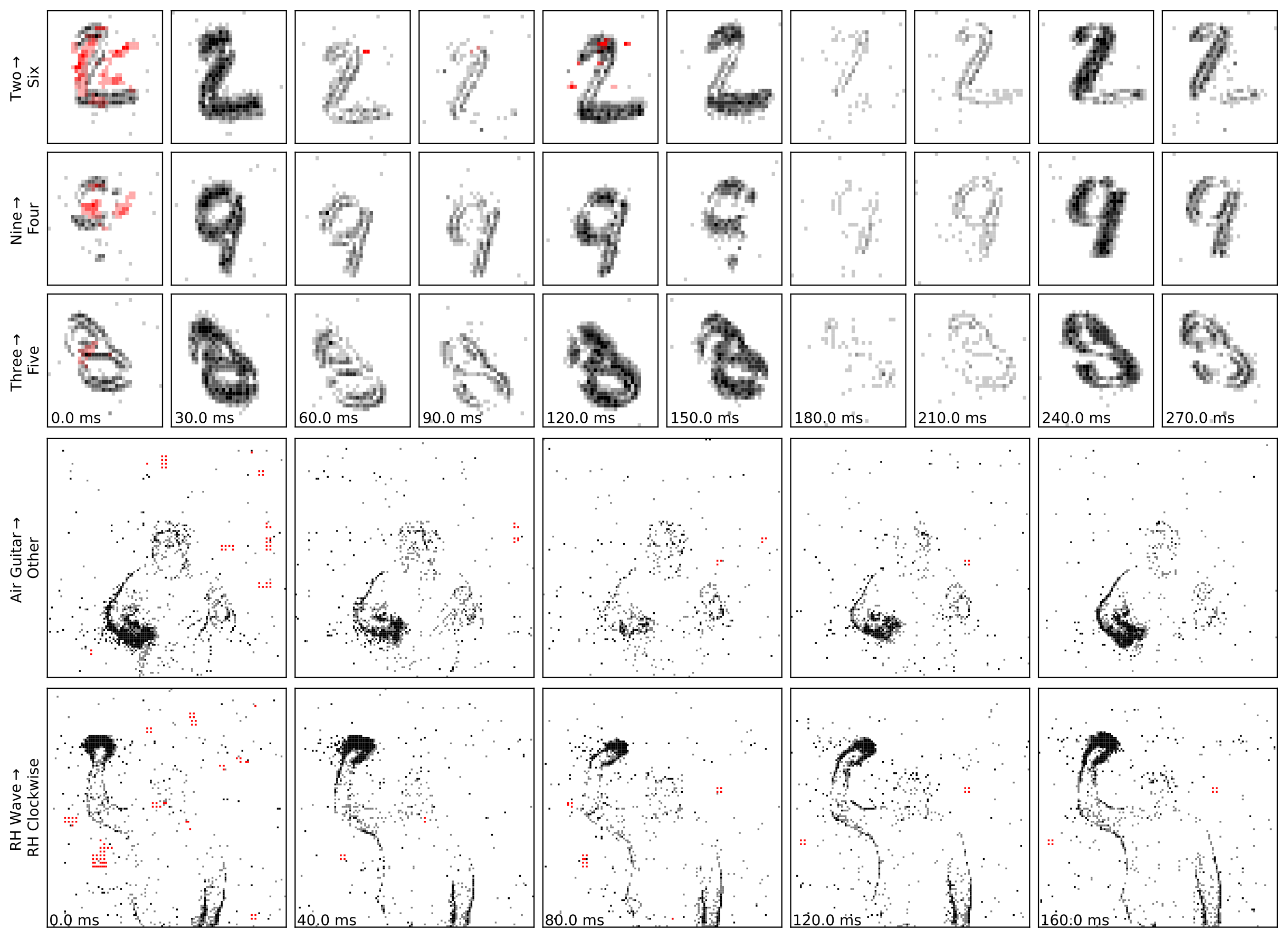} \\
    \caption{Examples of adversarial inputs on the, \gls{nmnist} (top) and IBM Gestures (bottom) datasets, as obtained by the SpikeFool method. The captions show the original (true) label, correctly identified, and the class later identified by the model. The data was re-framed in time for convenience of visualization. Red indicates added spikes. See the supplementary video for more examples and motion visualization.}
    \label{fig:samples}
\end{figure*}

\begin{figure*}
    \centering\includegraphics[width=.9\textwidth]{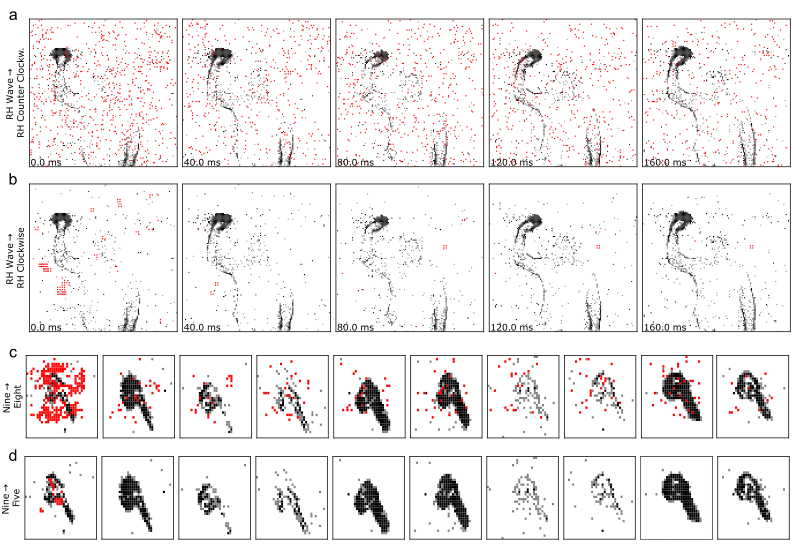} \\
    \caption{Our method (b,d) significantly reduces the magnitude of the perturbations introduced by the adversary. The runner-up method proposed in \cite{adv-snn} (a,c) adds visibly more events and is thus less stealthy.}
    \label{fig:liang_comparison}
\end{figure*}

\subsection{Adversarial attack performance}
Table \ref{tab:comparisons} shows the comparison of SpikeFool with current state-of-the-art attack algorithms in the spiking domain on two widely used benchmarks. We compare various metrics such as success rate, median elapsed time per sample, and, most importantly, the median perturbation size. Results from \cite{marchisio2021dvs} and \cite{adv-snn} are drawn from our own implementations, closely following published code (if available) from the authors.

\begin{table*}[h]
    \centering
        \resizebox{0.8\textwidth}{!}{%
        \begin{tabular}{ccccccc}
        \toprule
        & Attack strategy & Hardware & \makecell{Success\\Rate (\%)} & \makecell{Median Elapsed\\Time (s/sample)} & \makecell{Median \\No. Queries} & \makecell{Median\\$L^0$}\\
        \midrule
        \multirow{4}{*}{\rotatebox[origin=c]{90}{N-MNIST}}
        & SpikeFool ($\eta=0.2,\lambda=2$) & GPU & 99.53 & $24.87\pm 0.0815$ & 45 & $\mathbf{256}$\\
        & SpikeFool ($\eta=0.5,\lambda=2$) & GPU &  99.88 & $12.00\pm 0.072$ & 27 & 272 \\
        & \citealt{adv-snn} & GPU & $\mathbf{100.00 \pm 0.00}$ & $0.69\pm 1e-4$ & $\mathbf{2}$ & $865.3 \pm 6.37$ \\
        & \citealt{marchisio2021dvs} & GPU &  $74.41\pm 0.33$ & $\mathbf{0.64\pm 0.001}$ & 3 & $3.23e4 \pm 360.32$ \\
        \midrule
        \multirow{5}{*}{\rotatebox[origin=c]{90}{IBM}}
        & SpikeFool ($\eta=0.1,\lambda=3$) & GPU &  $\mathbf{100.00}$ & $2.56\pm 0.003$ & 11 & 311 \\
        & SpikeFool ($\eta=0.1,\lambda=2$) & GPU &  99.87 & $2.4 \pm 0.002$ & 11 & 200\\
        & SpikeFool ($\eta=0.1,\lambda=1$) & GPU &  97.61 & $2.85\pm0.02$ & 17 & $\mathbf{116}$\\
        & SpikeFool & Speck2b &  88.05 & &  & \\
        & \citealt{adv-snn} & GPU &  $99.77 \pm 0.126$ & $0.53 \pm 0.004$ & 9 & $1345.5 \pm 11.48$ \\
        & \citealt{marchisio2021dvs} & GPU &  $92.44 \pm 0.05$ & $\mathbf{0.14 \pm 2e-4}$ & $\mathbf{2}$ & $227343 \pm 54.56$ \\
        \bottomrule
        \end{tabular}%
        }
    \newline $^\dagger$ \footnotesize{Samples for which the attack was unsuccessful were considered to have undefined $L^0$. Attacks were repeated 5 times on 1000 samples. We report the mean and standard deviation for non-deterministic metrics. The median number of queries did not show any variation over 5 repetitions. \\ \vspace{0.3em}}
    \caption{Result comparison between SpikeFool and related works. 
    $\eta$ indicates the minimum step size for updates to the perturbation: higher values of $\eta$ find less precise perturbations (larger $L^0$ values), but are sometimes needed in order to prevent zero-gradient issues within the algorithm. $\lambda$ is the sparsity parameter: lower $\lambda$ (with a minimum of 1) yields sparser results, but gives a slightly lower success rate.
    The success rate is defined as the fraction of samples that were initially correctly classified, for which the attack algorithm converged to an adversarial example that the network classifies incorrectly.}
    \label{tab:comparisons}
\end{table*}

We find that the attack presented in \cite{marchisio2021dvs} largely fails to achieve reasonable perturbation magnitudes on both benchmarks. This is likely because of the lack of bounds and rounding in their method, which result in large deviations.
In contrast, the attack presented in \cite{adv-snn} achieves near perfect success rates in very short time at relatively low perturbation sizes. However, this speed comes at the cost of perturbation size: our method yields perturbations that are up to $11\times$ smaller than the ones generated by \cite{adv-snn}. To put these numbers into perspective, Figure \ref{fig:liang_comparison} compares the adversarial samples generated by both methods. One can observe the clear difference between the two methods and could argue that attacks generated by \cite{adv-snn} are more visible.
Interestingly, we observe that SpikeFool often resorts to inserting small patches limited to key areas. We believe this is due to the fact that the network is inherently robust to salt-and-pepper noise and that introducing localized patches is by far more effective. Figure \ref{fig:samples} and the supplementary video show additional examples of successful attacks. Further information about how many spikes are added and removed during training can be found in the supplementary material.

We also run a random subset of 777 DVS Gesture samples that were attacked using SpikeFool ($\lambda=2.0,\eta=0.3$) on our neuromorphic hardware and observe a success rate of 88.05\% misclassified samples, in comparison to 93.28\% in simulation. This performance is lower compared to the one reported in Table \ref{tab:comparisons} due to the much higher time resolution needed on chip, which makes it harder to find an attack.   
A possible reason for this discrepancy lies in how the chip is limited in computing capacity by weight precision and restricted throughput per time unit, which causes some of the input events to be dropped. Furthermore, the conversion of binned data back into lists of spikes is necessarily lossy.
In terms of attack efficiency, we observe a median difference in number of spikes of 903 among the attacks that were successful on chip, corresponding to a median 9.3\% increase in the number of spikes per sample.

\subsection{Adversarial patches}
Although we have demonstrated that one can achieve high success rates on custom spiking hardware that operates with microsecond precision, the applicability of this method is still limited, as the adversary needs to suppress and add events at high spatial and temporal resolution, thus making the assumption that the adversary can modify the event-stream coming from the \gls{dvs} camera. Furthermore, SpikeFool assumes knowledge of the model and requires computing the perturbation offline, which is not feasible in a timely manner. In a more realistic setting, the adversary is assumed to generate perturbations by changing the input the \gls{dvs} camera receives on the fly, by e.g. adding a light-emitting device to the visual input of the \gls{dvs} camera.

Using the training data from the IBM Gestures dataset, we generate an adversarial patch for each target class with high temporal precision (event samples of 200 ms are binned using 0.5 ms-wide bins) and evaluate the effectiveness in triggering a targeted misclassification both in simulation and on-chip using the test data. To simulate spatial imprecision during deployment, each test sample is perturbed by a patch that was randomly placed within the area of the original gesture.
Table \ref{tab:patches} summarizes our findings on target success rates for generated and random patches. 
Simulated results show high success rates, and on-chip performance shows a slight degradation, which can be expected due to weight quantization on the tested specialized hardware. 
We also find that the chip has trouble processing inputs because most of the added patch events occur concentrated in the beginning of recordings in a large transient peak. 
In one case, the targeted attack for label ``Arm Roll`` mostly fails on chip as not all events are processed, which makes it harder to differentiate from  ``Hand Clap'', a similar gesture that occurs in the same central spatial location. 
This could be mitigated by limiting the number of events in a patch to ensure that they could all be correctly processed on the chip. 

\begin{table*}
    \centering
        \resizebox{0.9\textwidth}{!}{%
        \begin{tabular}{ccccccccccccc}
        \toprule
        & Target label & \makecell{Hand \\clap} & \makecell{RH \\Wave} & \makecell{LH \\Wave} & \makecell{RH \\Clockwise}  & \makecell{RH Counter\\Clockwise} & \makecell{LH \\Clockwise}  & \makecell{LH Counter\\Clockwise} & \makecell{Arm \\Roll} & 
        \makecell{Air \\Drum} & \makecell{Air\\ Guitar} & \makecell{Other} \\
        \midrule
        \multirow{4}{*}
        & Adversarial patch & 90.3 & 99.0 & 89.8 & 87.3 & 79.7 & 49.7 & 51.5 & 63.6 & 79.1 & 92.3 & 64.7 \\
        & Adv. patch (on-chip) & 94.0 & 89.0 & 94.1 & 81.3 & 65.1 & 35.9 & 43.8 & 5.0 & 82.7 & 87.3 & 66.8 \\
        & Random patch & 18.8 & 80.7 & 77.0 & 0 & 0 & 3.6 & 0.6 & 0 & 0 & 12.6 & 16.6 \\
        & Rand. patch (on-chip) & 43 & 76.8 & 72.2 & 0 & 0 & 9.0 & 2.4 & 0 & 0 & 0 & 17.7 \\

        \bottomrule
        
        \end{tabular}%
        }
    \caption{Adversarial patches for different target labels were evaluated on-- and off--chip. Shown here are the success rates in percent for each target label. An attack is considered successful if the original label is not the target label and the network predicts the target label when the patch is applied.}
    \label{tab:patches}
\end{table*}

\begin{figure*}
    \centering\includegraphics[width=.9\textwidth]{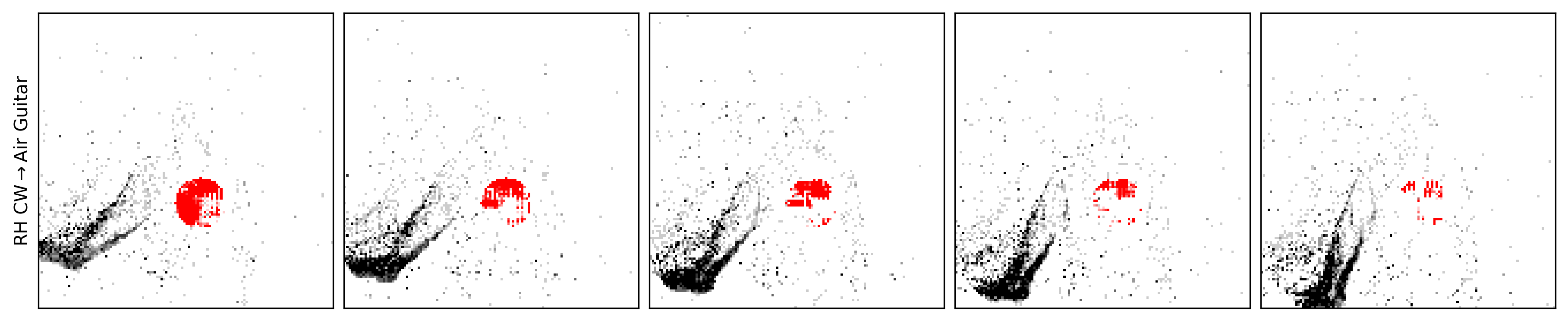}\vspace{-0.5em}
    \centering\includegraphics[width=.9\textwidth]{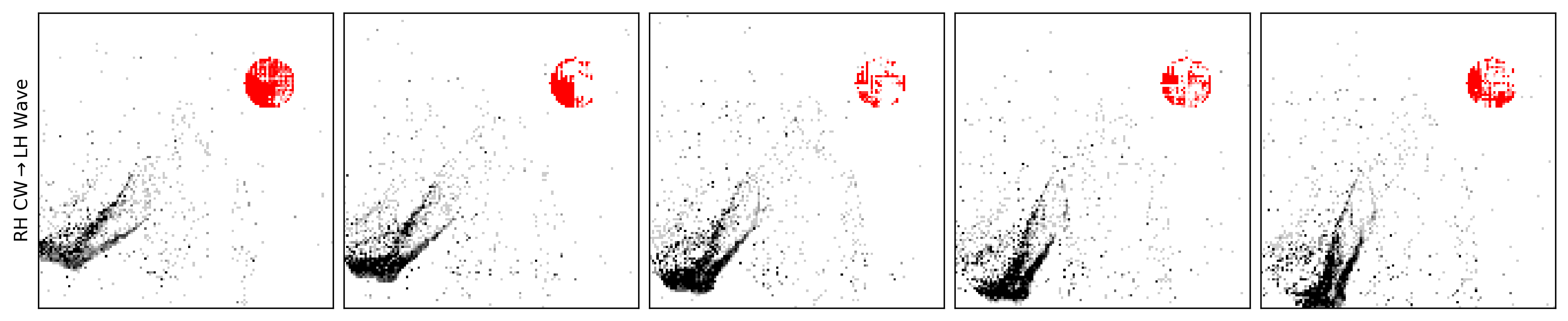}\vspace{-0.5em}
    \centering\includegraphics[width=.9\textwidth]{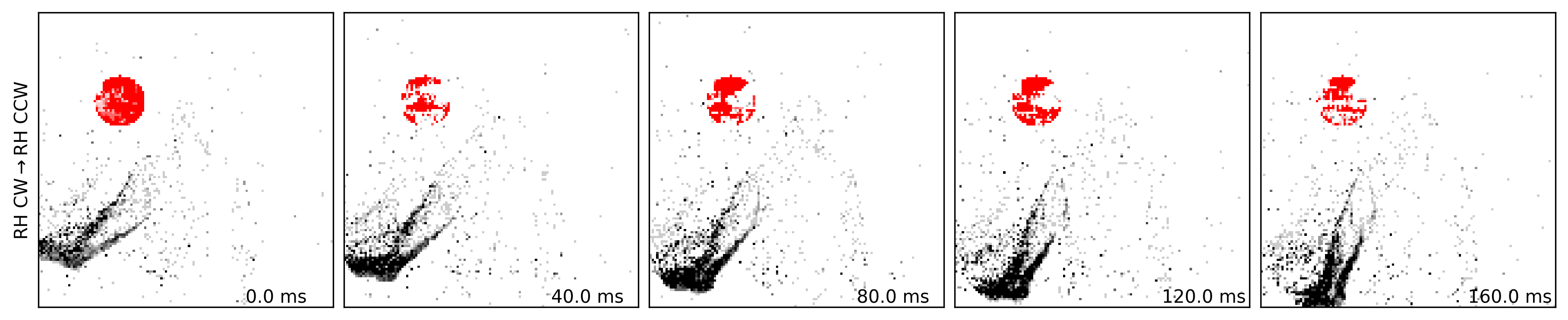} \\
    \caption{Examples of adversarial patches successfully applied to a single ``right hand clockwise'' data sample, with different target classes. See also the supplementary video for motion visualization and more examples.}
    \label{fig:samples_patches}
\end{figure*}

We compare this result with a baseline of randomly generated patches, and we observe that two labels, namely ``Left`` and ``Right Hand Wave'' subsume all other attacked labels in this case. This hints that randomly injecting events in various locations is not enough to perturb network prediction to a desired label and that our algorithm succeeds in finding a meaningful patch. 
We find that generating patches with lower temporal resolution heavily degrades performance on-chip, as the chip operates using microsecond precision. We also perform ablation studies where we move the patch far from the position where the gesture is expected and observe that the patch mostly triggers a different misclassification, mostly the gesture that is expected at the new position. We find that this observation originates from the fact that the network heavily relies on the spatial location of the gesture and one should, if the hardware allows, consider larger networks that are invariant to the position of the gesture. 
To summarize,  adversarial patches are effective in triggering a targeted misclassification both on-- and off--chip compared to randomly generated ones. Figure \ref{fig:samples_patches} and the supplementary video show examples of successful patch attacks. Importantly, these attacks are universal, meaning that they can be applied to any input and do not need to be generated for each sample.

\subsection{Defense via adversarial training}\label{sec:defense}
Once it is known that a model or system is sensitive to a certain type of adversarial attack, it is natural to investigate whether there is a way to build a network that is more resistant to these attacks. We therefore experimented with adversarial training using the TRadeoff-inspired Adversarial Defense via Surrogate-loss minimization (TRADES) method \citep{trades}. The method adds a new term to the loss function, which minimizes the Kullback-Leibler divergence between the output of the network on the original input and the output when the adversarial example is presented:
$$ \mathcal{L}_\text{rob} = \mathcal{L} + \frac{\beta_{\text{rob}}}{B}\operatorname{D_\text{KL}}(f(\mathbf{x}_\text{adv}); f(\mathbf{x}_0)).  $$
Here, $B$ is the batch size, $\beta_{\text{rob}}$ is the parameter that defines the trade-off between robustness and accuracy, $f$ is the network and $\mathbf{x}_\text{adv}$ is the adversarial input.
To find $\mathbf{x}_\text{adv}$ at training time, we use a \gls{pgd}-based attack, since it can be batched --- but we attack the resulting networks using SpikeFool at test time. 

More specifically, we use \gls{pgd} in the $L^\infty$ domain and choose $\epsilon=0.5$ as the maximum perturbation, with $N_\text{pgd}=5$ attack steps. We use \gls{pgd} in the spiking domain by accumulating the gradients in full-precision using a straight-through estimator~\citep{sfe}. The details of our spiking-adapted implementation of \gls{pgd} are described in the supplementary material, including extensions, and a full comparison with SpikeFool.

\begin{figure*}
    \centering
    \includegraphics[width=.95\textwidth]{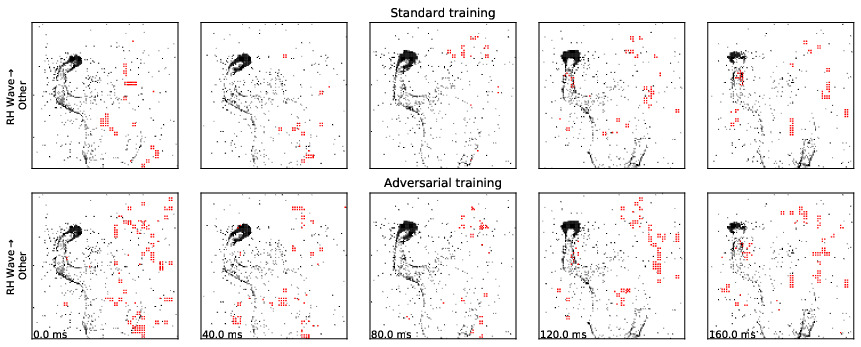}
    \caption{Adversarial training visibly increases the number of events that need to be added/removed in order to trigger a misclassification.}
  \label{fig:defense}
\end{figure*}

Although we use a much simpler attack strategy at training time, we found that it produced perturbations of reasonable sizes while being extremely efficient and sufficiently effective: for the choices of $\beta_{\text{rob}}$ we considered, we see that the adversarially-trained network requires stronger and less stealthy attacks before it is fooled (Figure \ref{fig:defense}). As expected, SparseFool's success rate is still high, since it aims to find a solution no matter the costs; but there is an increase in the number of added spikes required, which is indeed a sign of robustness (Table \ref{tab:defense}). Further work is required for a comprehensive investigation of other defense strategies.

\begin{table}
    \centering
        \resizebox{\columnwidth}{!}{%
        \begin{tabular}{ccccc}
        \toprule
        & Network & \makecell{Test Accuracy (\%)} & \makecell{Success Rate (\%)} & \makecell{Median $L^0$}\\
        \midrule
        \multirow{5}{*}{\rotatebox[origin=c]{90}{IBM}}
        & $\beta_{\text{rob}}=0.0$ (normal) & 79.23 & 98.19 & 438.0\\
        & $\beta_{\text{rob}}=0.01$ & \textbf{81.36} & \textbf{98.78} & 511.0\\
        & $\beta_{\text{rob}}=0.05$ & 75.27 & 97.65 & \textbf{1434.0}\\
        & $\beta_{\text{rob}}=0.1$ & 73.16 & 95.46 & 1316.0\\
        \bottomrule
        \end{tabular}%
        }
    \caption{Training with a small robustness term ($\beta_{\text{rob}}=0.01$) increases generalization and therefore improves the test accuracy of the network. For increasing values of $\beta_{\text{rob}}$, the amount of changed events needed to trigger a misclassification increases dramatically (more than $3\times$), while the performance of the network decreases (hence the name ``tradeoff-inspired'').}
    \label{tab:defense}
\end{table}

\section*{Discussion}
We studied the possibility of fooling \glspl{snn} through adversarial perturbations to \gls{dvs} data, and verified these perturbations on a spiking convolutional neuromorphic chip. There were two main challenges to this endeavor: the discrete nature of event-based data, and their dependence on time. \gls{dvs} attacks also have different sparsity requirements, because the magnitude of the perturbation is measured in terms of number of events added or removed. For this purpose, we adopted a surrogate-gradient method and backpropagation-through-time to perform white-box attacks on spiking networks. We presented SpikeFool, and adapted version of SparseFool, which we compared to current state-of-the-art methods on well-known benchmarks. We find that SpikeFool achieves near perfect success rates at lowest perturbation magnitudes on time-discretized samples of the \gls{nmnist} and IBM Gestures datasets. In the best cases, the attack requires the addition of less than a hundred events over 200 ms. To the best of our knowledge, we were also the first to show that the perturbation is effective on a network deployed on a neuromorphic chip, implying that the method is resilient to the small but non-trivial mismatch between simulated and deployed networks.

Additionally, since SpikeFool computes perturbations offline and not on a live stream of \gls{dvs} events, we also investigated a more realistic setting, where an adversary can inject spurious events in the form of a patch inserted into the visual field of the \gls{dvs} camera. We demonstrated that we can generate patches for different target labels. Although these patches require a much higher amount of added events, they do not require prior knowledge of the input sample and therefore offer a realistic way of fooling deployed neuromorphic systems.
A natural next step would be to understand whether it is possible to build real-world patches that can fool the system from a variety of distances and orientations, as \cite{eykholt2018robust} did for photographs. Moreover, it will be interesting to see how important knowledge about the architecture is and if one can generate patches by having access to a network that differs from the one deployed.


\section*{Conflict of Interest Statement}
Author J.B. is employed by IBM. Authors G.L., Y.H. and S.S. are employed by SynSense. The remaining authors declare that the research was conducted in the absence of any commercial or financial relationships that could be construed as a potential conflict of interest.

\section*{Author Contributions}
J.B. and M.S. conceived and designed the research. All authors developed software and simulations. J.B., M.S., G.L. and Y.H. performed experiments and collected data. J.B., M.S., G.L. and Y.H. analyzed and interpreted the data. J.B., M.S. and G.L. drafted the manuscript. All authors performed critical revision of the manuscript and approved the final version for publication.

\section*{Funding}
M.S. was supported by an ETH AI Center postdoctoral fellowship during part of this work.

\section*{Acknowledgments}
The authors would like to thank Seyed Moosavi-Dezfooli for valuable help in understanding and applying the SparseFool and DeepFool algorithms, and the Algorithms team at SynSense for their support, suggestions, and assistance in using the Speck chip. We also thank the anonymous reviewers for many helpful comments.

\section*{Data Availability Statement}
The code and links to the datasets used for running the experiments are publicly available at \url{https://github.com/jubueche/SPAA}

\bibliography{bibliography}

\newpage
\section{PGD adapted to the spiking domain}
\subsection{Continuous-discrete PGD} \label{sec:cdpgd}
\gls{pgd} \citep{madry} is a standard attack algorithm that iteratively computes the gradient of a loss function w.r.t. the inputs, takes a step in the direction of the gradient and, if needed, reduces the perturbation to fulfill a constraint on the maximum perturbation magnitude. For this algorithm to work in the spiking domain, some changes are necessary. The first modification is given by rounding the adversarial input after every update. However, doing so, updates are retained only if the gradient magnitude is large enough, otherwise the small changes made to the input are lost due to the subsequent discretization. Instead, we use straight-through estimation~\citep{sfe}, an approach that prevents this loss of information: we keep a \textit{continuous} version of the image as a copy, but use the gradients computed on the \textit{discretized} image to update the continuous version which is kept in memory. This lets us accumulate updates across iterations independently of rounding.

Secondly, after convergence, to adapt \gls{pgd} to the scenario where we want to find the smallest perturbation that triggers a misclassification, we sort the continuous values based on the difference between the original data and the continuous perturbed version. We then iterate through the sorted list of indices and create a final discrete data sample starting with the original input and flipping each binary value in order, until a misclassification is triggered.
The latter process improves the perturbation size, but comes at a high computational cost. Therefore, we chose to skip this step when using continuous-discrete PGD for adversarial training (Section \ref{sec:defense} of the main text).

\subsection{Probabilistic PGD}
We also devised an alternative way of using \gls{pgd} on discrete data, which we call ``Probabilistic \gls{pgd}". Probabilistic \gls{pgd} works by assuming that the binary input was generated by sampling from a series of independent Bernoulli random variables. This approach aligns with how the \gls{dvs} camera generates thexamplese binary data: the probability of emitting a spike at time $t$ is proportional to the change in light intensity, a continuous metric. For each round of \gls{pgd}, the input is sampled in a differentiable manner by the Gumbel-softmax reparameterization trick \citep{gumbel}:

\begin{equation*}
    \mathbf{x_{\text{adv}}} = \sigma\left(\frac{\text{log}(\mathbf{r})-\text{log}(\mathbf{1}-\mathbf{r}) + \text{log}(\mathbf{p}_{\text{adv}}) - \text{log}(\mathbf{1}-\mathbf{p}_{\text{adv}}) }{T}\right)
\end{equation*}
where $\mathbf{r} \sim \mathcal{U}(\mathbf{0},\mathbf{1})$, $T=0.01$ is a temperature parameter, and $\sigma$ is the sigmoid function. Note that bold-faced variables indicate matrices in the shape of the input to the neural network. The underlying probabilities $\mathbf{p}_{\text{adv}}$, instead of the pixel values $\mathbf{x_{\text{adv}}}$, are updated using the gradient obtained from the loss function that is minimized by \gls{pgd}. We observed that this generally improved performance compared to the \gls{pgd} version explained above (Table \ref{tab:comparisons-sm}). Gradients are averaged over $N_{\text{mc}}=10$ samples of $\mathbf{r}$. It should be noted that the need for a gradient sampling procedure significantly increases the runtime.

\subsection{Comparison to SpikeFool}
Table \ref{tab:comparisons-sm} contains the results of our comparison analysis.

\begin{table*}
    \centering
        \resizebox{0.8\textwidth}{!}{%
        \begin{tabular}{cccccc}
        \toprule
        & Attack strategy & \makecell{Success\\Rate (\%)} & \makecell{Median Elapsed\\Time (s/sample)} & \makecell{Median \\No. Queries} & \makecell{Median\\$L^0$}\\
        \midrule
        \multirow{4}{*}{\rotatebox[origin=c]{90}{N-MNIST}}
        & Continuous-discrete \gls{pgd} & 48.63 & 72.56 & 1052 & $^\dagger$\\
        & Probabilistic \gls{pgd} & \textbf{100.00} & 88.99 & 1091 & 839\\
        & SpikeFool ($\eta=0.2,\lambda=2$) & 99.76 & 30.22 & 45 & \textbf{254}\\
        & SpikeFool ($\eta=0.5,\lambda=2$) &  99.88 & \textbf{13.08} & \textbf{26} & 268 \\
        \midrule
        \multirow{5}{*}{\rotatebox[origin=c]{90}{IBM}}
        & Continuous-discrete \gls{pgd} & 88.30 & 16.68 & 747 & 695\\
        & Probabilistic \gls{pgd} & 99.22 & 16.80 & 555 & 303\\
        & SpikeFool ($\eta=0.1,\lambda=3$) &  \textbf{100.00} & 2.78 & 11 & 310 \\
        & SpikeFool ($\eta=0.1,\lambda=2$) &  99.87 & \textbf{2.57} & 11 & 200\\
        & SpikeFool ($\eta=0.1,\lambda=1$) &  97.69 & 3.02 & 17 & \textbf{116}\\
        \bottomrule
        \end{tabular}%
        }
    \newline $^\dagger$\small{Samples for which the attack was unsuccessful were considered to have undefined $L^0$. Because C-D \gls{pgd} fails more than half of the time, the median is undefined.\\ \vspace{0.3em}}
    \caption{Result comparison between \gls{pgd}-based methods and SpikeFool.}
    \label{tab:comparisons-sm}
\end{table*}

\section{The binarized MNIST dataset} \label{sec:bmnist}

\begin{table}[b]
    \centering
        \resizebox{1\columnwidth}{!}{%
        \begin{tabular}{cccccc}
        \toprule
        & Attack & \makecell{Success\\Rate (\%)} & \makecell{Median Elapsed\\Time (s/sample)} & \makecell{Median \\No. Queries} & \makecell{Median\\$L^0$}\\
        \midrule
        \multirow{4}{*}{\rotatebox[origin=c]{90}{B-MNIST}}
        & SCAR & \textbf{100.00} & 1.14 & 1175 & \textbf{7}\\
        & Continuous-discrete PGD & 98.89 & 0.16 & 102 & 50\\
        & Probabilistic PGD & 99.70 & 0.54 & 275 & 23\\
        & SpikeFool ($\eta=0.2,\lambda=2$) & 99.90 & \textbf{0.08} & \textbf{11} & 14\\
        \bottomrule
        \end{tabular}%
        }
    \caption{Comparison of attack strategies for B-MNIST. SCAR was implemented according to the pseudocode in~\cite{scar}.}
    \label{tab:bmnist}
\end{table}

\begin{figure*}
    \centering
    \includegraphics[width=0.9\textwidth]{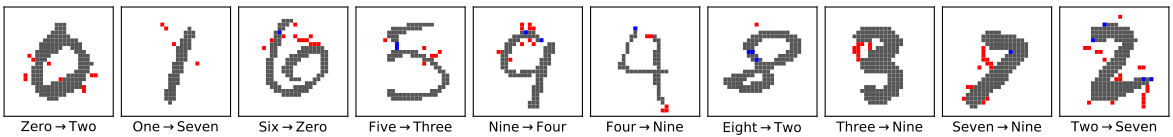}
    \caption{Example of adversarial attacks on B-MNIST. Blue indicates removed pixels, red added pixels. We note that in this lower-dimensional case, the effect of the attack is semantically interpretable: for example, adding a stroke that closes the upper left part of a ``7" makes it look like a ``9" not only for the network but also for a human observer.}
  \label{fig:bmnist}
\end{figure*}

We tried our methods on an additional dataset. This is a binarized version of MNIST (B-MNIST for short), which is derived from  the popular MNIST Handwritten Digits database \citep{mnist}, binarized so that pixel values 0 to 127 are mapped to white, and 128 to 255 are mapped to black. No other preprocessing is applied. This is \emph{not} a dataset of DVS recordings: we use it in order to compare our white box attacks against the SCAR attacks for binary datasets \citep{scar}.  SCAR is a black-box algorithm, i.e.\ it only assumes access to the output probabilities of the network. The algorithm flips bits in areas chosen according to a specific heuristic and keeps flipped those that cause
a change in the confidence of the network.

\subsection{Results} Table \ref{tab:bmnist} compares the different algorithms on B-MNIST and shows that SpikeFool finds successful adversarial examples with a low median $L^0$ (i.e.~number of perturbed pixels), while requiring a very low median execution time. Figure \ref{fig:bmnist} illustrates samples of perturbations found by SpikeFool and the corresponding label that was predicted by the network after applying the perturbation. Because of the small sample size and the fact that there is no time dimension, B-MNIST enables us to compare SpikeFool to a less computationally efficient methods like SCAR. However, more realistic datasets are needed to truly evaluate the feasibility of applying these algorithms, as shown in the main text.

\subsection{Network} For the B-MNIST experiments, we use a non-spiking network, similar to the one used in \citep{scar}: two $3\times3$ convolutional layers (32 and 64 channels each), with ReLU activations, followed by $2\times2$ max-pooling, dropout, and a fully connected layer of 128 features, projecting onto the final layer of 10 output units. The network is trained for 50 epochs at batch size 64, using the Adam~\citep{adam} optimizer with learning rate $10^{-3}$ on a cross-entropy loss function. The network reached a test accuracy of 99.12\%.

\section{Empirical analysis of SpikeFool perturbations} \label{sec:analysis}

With the aim of gaining more insight into the behavior of our methods, we studied the characteristics of the perturbations resulting from SpikeFool attacks in more detail. For this, we focused on two specific experiments: a SpikeFool run on N-MNIST with hyperparameters $\eta = 0.5$ and $\lambda=2$; and the same IBM Gestures experiment that was run on-chip.

\begin{figure*}
    \centering
    \includegraphics[width=\textwidth]{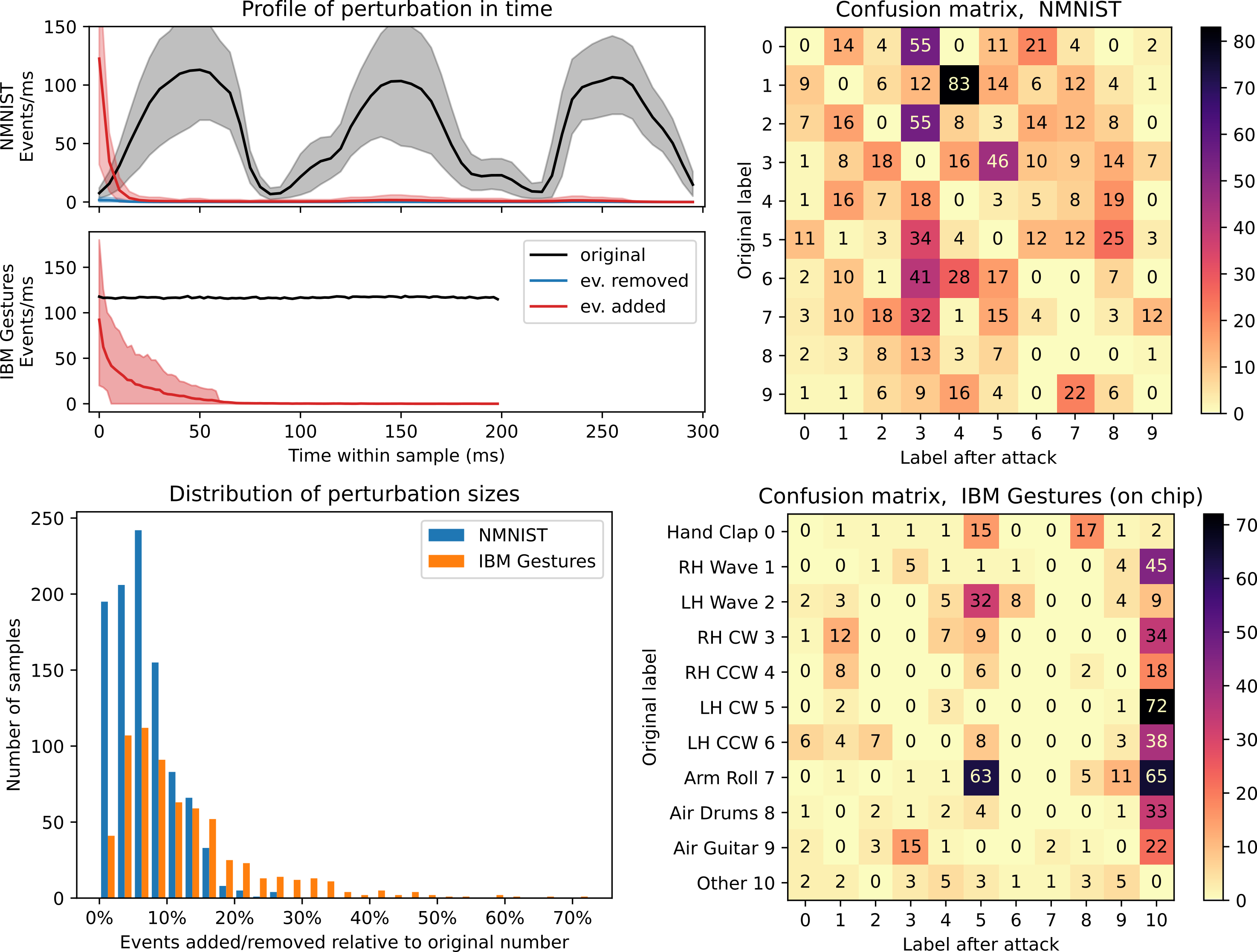}
    \caption{Properties of the adversarial perturbations found by SpikeFool, for two experiments: N-MNIST (in simulation, $\eta = 0.5, \lambda=2$) and IBM Gestures (as tested on chip). \textit{Top left}: Number of events in time within each data sample. The shaded areas represent the 0.1-0.9 inter-quantile range (not shown for the `original' curve in the bottom panel). The perturbation tends to consist of spikes added at the beginning of the sample, especially for N-MNIST which does not rely on temporal structure for inference. Very few spikes are removed, which justifies the choice of ignoring removed spikes in on-chip experiments. The periodic structure of N-MNIST samples is intrinsic to the dataset, recorded with saccades. \textit{Bottom left}: Distribution of increase in number of spikes after the attack, relative to the original number. \textit{Right}: Matrices showing the label identified by the network when presented with the adversarial examples, given the original label, for the two experiments. Most IBM Gestures classes are perturbed towards the `other' class, while there is no clear structure in the N-MNIST case. LH = Left Hand, RH = Right Hand, (C)CW = (Counter) ClockWise.}
  \label{fig:speck_sparsefool}
\end{figure*}

First, we notice that SpikeFool-based perturbations rarely involve the removal of events. In the N-MNIST experiment considered here, an average of 7.6 events are removed from each sample, compared to an average of 214 spikes added. This justified our choice to ignore removed events in the course of the on-chip experiments.
As is evident from the examples in the manuscript, we also find that SpikeFool's adversarial perturbations tend to insert spikes at the beginning of the sample, with only a few spikes added later in time. The top left panel of figure \ref{fig:speck_sparsefool} shows the time profile of the perturbations in detail. We believe this is a consequence of the use of the non-leaky neuron model. In non-leaky neurons, information can be stored indefinitely in the membrane potential, so early spikes have a bigger chance of contributing to a spike later in time, and are more effective compared to events added later in the sample. This effect is also present in the IBM Gestures experiment, but looks less prominent, possibly because networks trained with BPTT on data with richer features in time have non-trivial dynamics. In this sense, we expect this phenomenon to be further reduced or disappear entirely when the task is strictly linked to the time evolution of the input signal, such as in auditory speech recognition. The timing of adversarial events could potentially be used for model interpretability purposes, to measure how much the model relies on temporal features.

Further to the median values reported in table 1 of the main text, the lower left panel of figure \ref{fig:speck_sparsefool} reports the full distributions of the number of added or removed events. Here, we display the numbers relative to the original number of events in the sample. We notice a minority of cases where the attack is successful only at the cost of a very significant injection of events.

Finally, we analyzed the statistics of classes identified by the networks after the attack. SpikeFool is used as an ``untargeted'' algorithm, i.e. it attempts to change the output of the network but without requirements on what the new class should be. Unsurprisingly, the ``other gesture'' class is a natural target class for many ground truth classes, but there are some exceptions which we find rather natural, such as ``left hand wave'' gestures being most often converted to ``left hand clockwise''. Conversely, we observe no dominant target class in the N-MNIST experiment. If the target class structure is undesirable, targeted attacks can be used instead.

\section{Training details}
\subsection{NMNIST model}
The analog \gls{cnn} was trained with Adam at batch size 64 with learning rate $10^{-3}$ for 10 epochs. We then rescaled the weights by layer-wise global factors so that the 99th percentile of activity was the same at each layer, as described by \cite{rueckauer2017conversion}.

\end{document}